\documentclass[sigconf, nonacm]{acmart}
\usepackage{color}
\usepackage{colortbl}
\usepackage{bbm}
\usepackage[abs]{overpic}
\usepackage{color}
\definecolor{lvyellow}{rgb}{1,1,0.6}
\definecolor{lvred}{rgb}{1,0.6,0.6}
\definecolor{lvorange}{rgb}{1,0.8,0.6}

\def\BibTeX{{\rm B\kern-.05em{\sc i\kern-.025em b}\kern-.08emT\kern-.1667em\lower.7ex\hbox{E}\kern-.125emX}}

\begin{document}	
\title{Fast Enhancement for Non-Uniform Illumination Images using Light-weight CNNs}
	
\author{
	Feifan Lv$^{1}$,
	Bo Liu$^{1}$,
	Feng Lu$^{1,2,*}$
}
\affiliation{
\institution{$^{1}$State Key Laboratory of VR Technology and Systems, School of CSE, Beihang University, Beijing, China}
\institution{$^{2}$Peng Cheng Laboratory, Shenzhen, China}
\vspace{0.5cm}
}

\renewcommand{\shortauthors}{Feifan Lv, et al.}

\begin{abstract}
	This paper proposes a new light-weight convolutional neural network ($\approx5$k params) for non-uniform illumination image enhancement to handle color, exposure, contrast, noise and artifacts, etc., simultaneously and effectively. 
	More concretely, the input image is first enhanced using Retinex model from dual different aspects (enhancing under-exposure and suppressing over-exposure), respectively. Then, these two enhanced results and the original image are fused to obtain an image with satisfactory brightness, contrast and details. Finally, the extra noise and compression artifacts are removed to get the final result. 
	To train this network, we propose a semi-supervised retouching solution and construct a new dataset ($\approx82$k images) contains various scenes and light conditions. 
	Our model can enhance 0.5 mega-pixel (like 600$\times$800) images in real-time ($\approx50$ fps), which is faster than existing enhancement methods. Extensive experiments show that our solution is fast and effective to deal with non-uniform illumination images.
\end{abstract}


	
\newcommand{\etal}{\emph{et al.}}

\maketitle
\section{Introduction}
\begin{figure}[t]
	\centering
	\includegraphics[width=\linewidth]{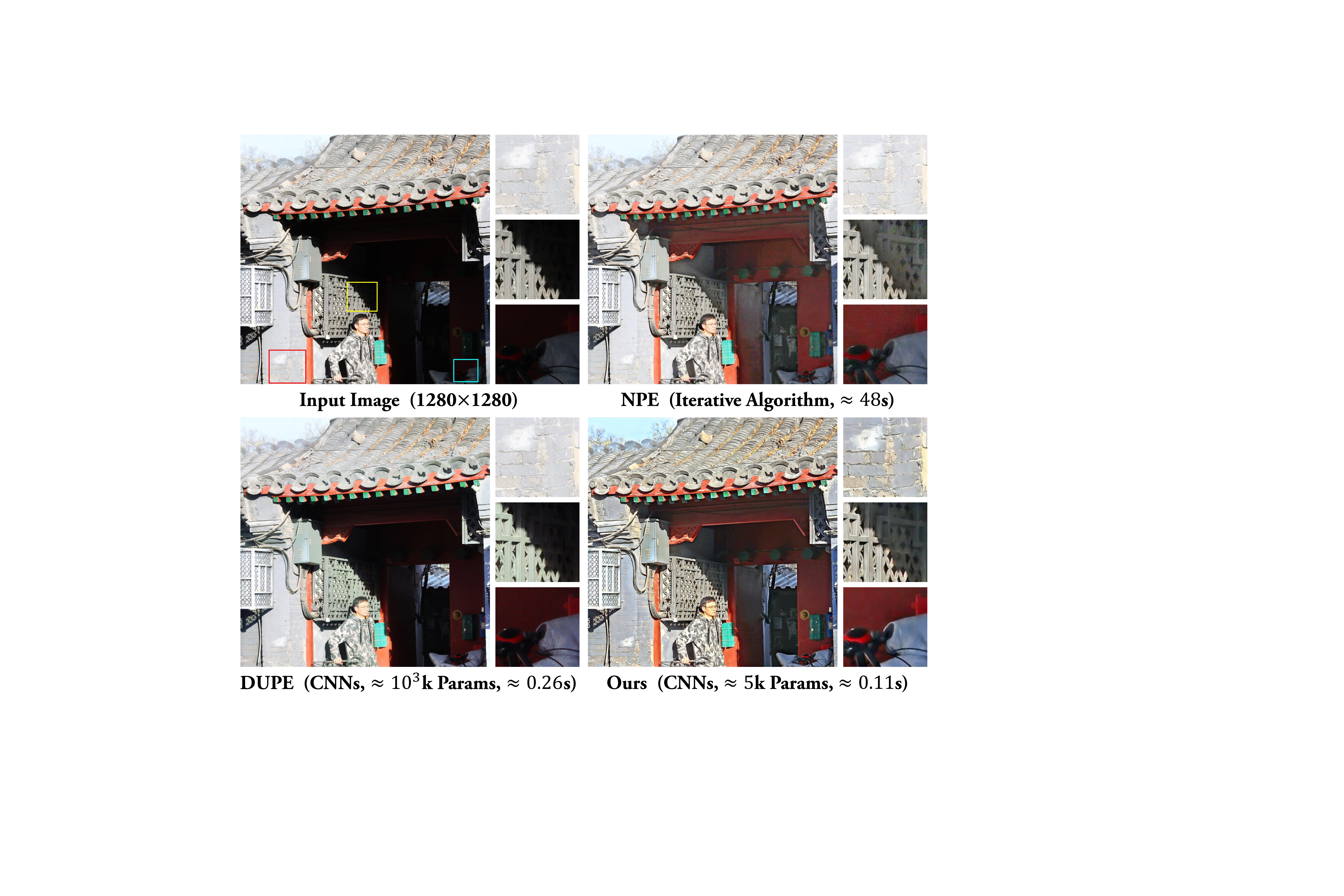}
	\caption{A challenging non-uniform illumination enhancement example. Comparing with existing methods, our solution can generate results with satisfactory visibility, vivid color, richer details and higher contrast using less time.}
	\label{sample}
\end{figure}

Due to the limitation of cameras' dynamic range and illumination, the photos we captured are usually with unsatisfactory visibility, dull colors, flat contrast and poor details, etc. This is especially noticeable in non-uniform illumination scenes, as shown in Figure~\ref{sample}. Fast enhancement for non-uniform illumination images thus will not only improve the visual quality of digital photography but also provide enough details for fundamental computer vision tasks, such as segmentation, detection and tracking, etc.

Non-uniform illumination image enhancement is a challenging task, as it needs to simultaneously manipulate many factors, such as color, contrast, exposure, noise, artifacts and so on. In addition, with the popularity of various camera sensors, like smartphone cameras, surveillance cameras, etc., the enhancement algorithms should be more light-weight and efficient to be applied for mobile devices and embedded systems.

Although many methods have been proposed to tackle this task in recent years, there is still large room for improvement whether in terms of performance or effect, as shown in Figure~\ref{sample}. 
Histogram equalization (HE) and Retinex theory~\cite{land1977retinex} are two typical traditional enhancement methods. 
HE-based algorithms~\cite{ibrahim2007bpdhe,arici2009wahe,nakai2013dheci,celik2011contextual,lee2013contrast} focus on improving the global contrast by stretching the dynamic range of images, which will result in limited local details and unnatural color. Retinex-based methods~\cite{jobson1997multiscale,lee2013amsr,fu2016mf,guo2017lime,li2018structure} try to recover the contrast by using the estimated illumination map. Mostly, they focus on restoring brightness and contrast while ignoring the influences of noise and artifacts. Learning-based methods~\cite{lore2017llnet,lvmbllen,gharbi2017deep,Chen2018Retinex,ren2019low} usually utilize heavy-weight and complex network architecture to deal with brightness, contrast, color and noise, which are difficult to apply to some real-time scenes or mobile devices. Besides, Learning-based methods need large images for training and the performance is limited by the quality of training dataset.

Therefore, in this paper, we first propose a novel semi-supervised pipeline to construct a paired image dataset for non-uniform illumination enhancement. Following the above pipeline, we build a paired dataset based on Microsoft COCO dataset~\cite{lin2014microsoft}, which contains numerous real-world image pairs with various exposure conditions. This dataset can be an efficient benchmark for enhancement researches.
Based on this dataset, we design a novel network for non-uniform illumination enhancement. In detail, it first enhances the non-uniform illumination images from both under- and over-expose aspects based on the Retinex model. Then, the different enhanced intermediate results are fused to generate the exposure corrected result. After that, the extra noise and compression artifacts are removed to get the final result. 
Our model is more light-weight ($\approx5$k parameters) and faster (enhance 0.5 mega-pixel images in real-time) than existing enhancement methods. 
Comprehensive experiments demonstrate that our method is superior to state-of-the-art methods in both qualitative and quantitative.

Overall, our contributions are in three folds:
\begin{itemize}
	\item We propose a novel light-weight network for non-uniform illumination enhancement, which can enhance images in real-time. It not only keeps the advantages of robustness of Retinex model but also overcomes the limitation of unable to enhance under-/over-exposure regions simultaneously.
	\item We construct a new large-scale dataset ($\approx82$k image pairs) for non-uniform illumination enhancement benchmarking and researching.
	\item Comprehensive experiments have been conducted to demonstrate that our method outperforms state-of-the-art methods qualitatively and quantitatively.
\end{itemize}

\section{Related Work}

Image enhancement has been studied and developed for a long time. In this section, we will make a brief overview of the most related methods.

{\bf Traditional enhancement methods.} Histogram equalization (HE) is a widely used technique by redistributing the luminous intensity on histogram. A lot of HE-based methods are proposed using additional priors and constraints. BPDHE~\cite{ibrahim2007bpdhe} preserves the mean brightness of the image to avoid unnecessary visual deterioration; Arici~\textit{et al}.~\cite{arici2009wahe} regards enhancement as an optimization problem and introduces specifically designed penalty terms; DHECI~\cite{nakai2013dheci} 
utilizes differential gray-levels histogram that contains edge information. These methods, however, focus on improving the contrast of the entire image without considering the illumination. Therefore, over- and under-enhancement often occur after adjustment.

Retinex theory~\cite{land1977retinex} supposes that an image is composed of reflection and illumination. Thus, MSR~\cite{jobson1997multiscale} and SSR~\cite{jobson1997properties}, recover and make use of the illumination map for low-light image enhancement. Furthermore, NPE~\cite{wang2013naturalness} makes a balance between details and naturalness. MF~\cite{fu2016mf} proposes a fusion-based method for weak illumination images. LIME~\cite{guo2017lime} develops a structure-aware smoothing model to improve the illumination consistency. BIMEF~\cite{ying2017bio} designs a multi-exposure fusion framework, and Ying~\textit{et al}.~\cite{ying2017newiccv} combine the camera response
model and traditional Retinex model. Mading~\textit{et al}.~\cite{li2018structure} consider a noise map for enhancing low-light images accompanied by intensive noise. However, most methods rely on hand-crafted illumination map and careful parameter tuning while can not deal well with noise and artifacts.

{\bf Learning-based enhancement methods.}
The past few years have witnessed the fast development of deep learning in the field of image enhancement. LLNet~\cite{lore2017llnet} trains a stacked sparse denoising autoencoder to learn the  brightening  and  denoising  functions. HDRNet~\cite{gharbi2017deep} designs an architecture to make local, global, and content-dependent decisions to approximate the desired image transformation. RetinexNet~\cite{Chen2018Retinex} combines the Retinex theory with CNN and KinD~\cite{zhang2019kindling} adds a Restoration-Net for noise removal. 
Wenqi~\textit{et al}.~\cite{ren2019low} use two distinct streams in hybrid network to simultaneously learn the global content and the salient structures.
DeepUPE~\cite{wang2019underexposed} introduces intermediate illumination in our network to associate the input with expected enhancement result, whereas it doesn't consider the noise in the low-light image. 
Besides, DPED~\cite{ignatov2017dslr} uses a  residual CNN to transform cameras from common smartphones into high-quality DSLR  cameras with the paired dataset. 
Differently, Yusheng~\textit{et al}.~\cite{chen2018deep} learn
image enhancement by GANs from a set of unpaired photographs with the user’s desired characteristics. 
As for extremely low-light scenes, SID~\cite{seedark2018cvpr} proposes a paired dataset and develops an end-to-end pipeline to directly process raw sensor images.
Most of these learning-based methods don't explicitly contain the denoising module, and some rely on traditional denoising methods with unsatisfactory results. What's more, these methods can not meet the real-time running demand for mobile devices.

Overall, the existing methods can hardly deal well with non-uniform illumination images both in quality and efficiency. In contrast, our approach is more light-weight and faster, and can enhance under-/over-exposure regions and restore the degradation simultaneously. Besides, our proposed dataset supplements non-uniform illumination enhancement benchmark datasets.
Therefore, our method is complementary to existing methods.

\section{Dataset}
In this section, we first compare the proposed dataset with existing enhancement datasets to demonstrate the reason of constructing a new dataset. After that, we introduce the construction details of our new dataset.

\begin{figure}[t]
	\centering
	\includegraphics[width=1\linewidth]{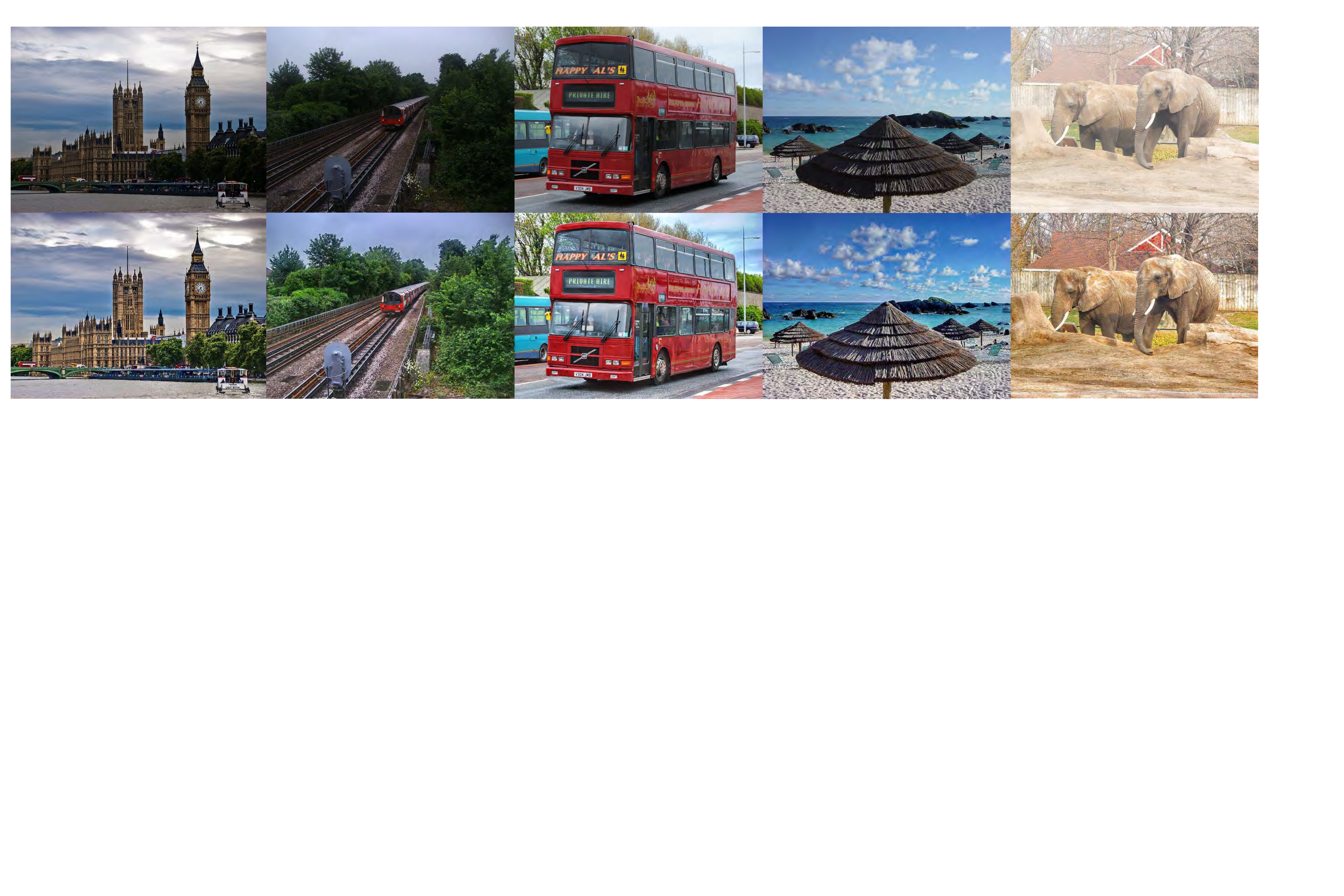}
	\caption{Example images of our dataset. {\bf Top:} non-uniform exposed images. {\bf Bottom:} corresponding reference images.}
	\label{fig:dataset}
\end{figure}

\begin{figure*}[htbp]
	\centering
	\includegraphics[width=1\textwidth]{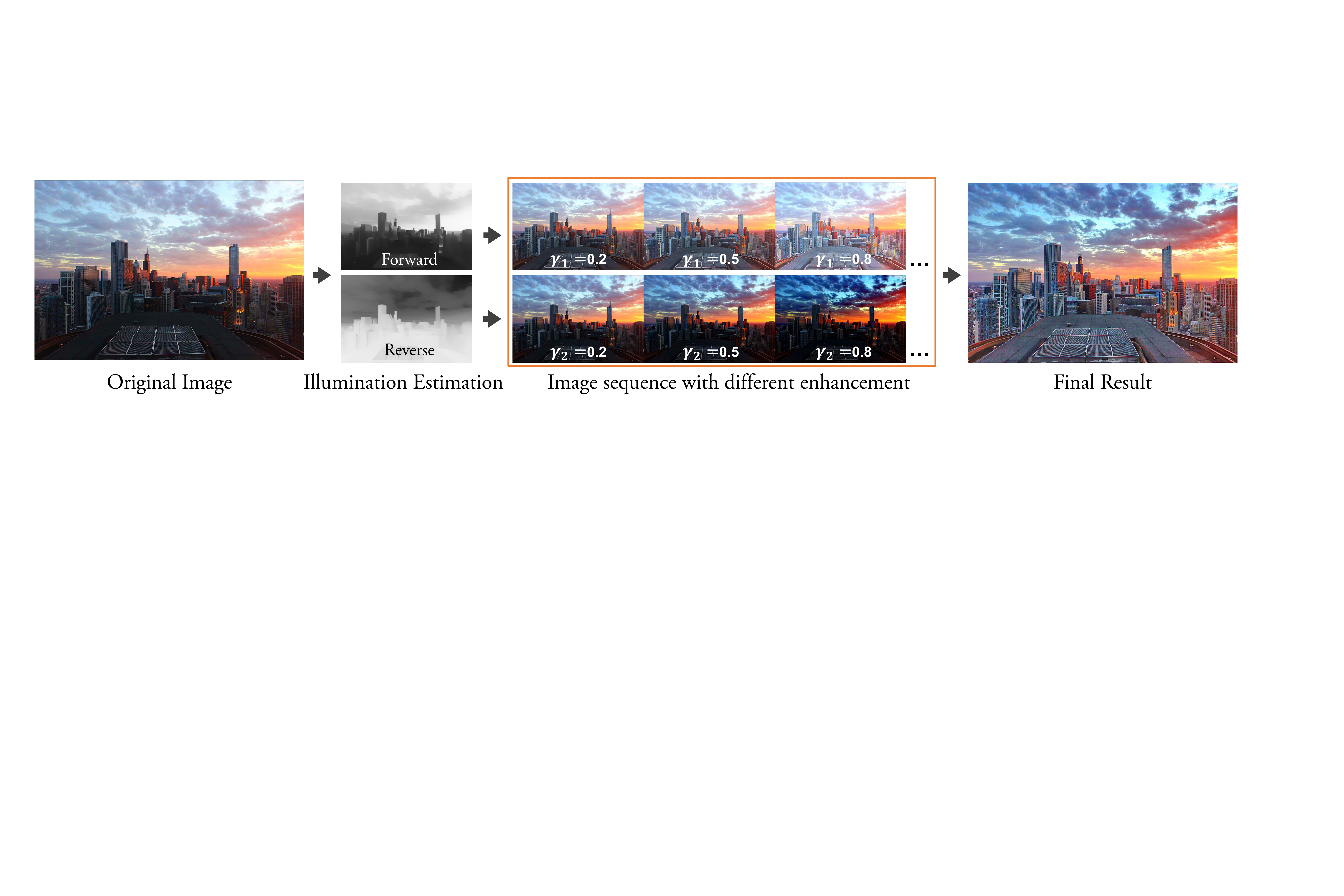}
	\caption{The pipeline of the proposed retouching module. We use the smoothness of bright channel (the maximal value of three channels) to replace the complex illumination estimation. Details can be found in Section~\ref{sec:dataset_details}.}
	\label{fig:simulation}
\end{figure*}

\subsection{Comparison with Existing Datasets}
There are two prevalent solutions to obtain paired differently exposed images: multiple shooting and expert retouching.
LOL~\cite{Chen2018Retinex} (altering ISO), SID~\cite{seedark2018cvpr} (altering exposure time) and DSLR~\cite{ignatov2017dslr} (altering hardware) are the representative datasets of the former solution. 
However, multiple shooting is time-consuming and labor-intensive, which limits the size of datasets, and faces the problem of image alignment. Besides, for the high-dynamic range scenes, even DSLR is difficult to get satisfactory results by shooting only once.  
To this dilemma, SICE~\cite{Cai2018deep} collects multi-exposure image sequences and uses Exposure Fusion techniques to construct the reference images, which are difficult to avoid the situation of blur and ghosting caused by incomplete alignment. DeepUPE~\cite{wang2019underexposed} and MIT-Adobe FiveK~\cite{bychkovsky2011learning} are the representative datasets of the latter solution, and are created for enhancing under-exposed and general images respectively. However, they lack the consideration of over-exposure scenes resulting in covering limited lighting conditions. To cover various lighting conditions and scenes, we take the expert retouching solution to construct a new dataset based on Microsoft COCO dataset~\cite{lin2014microsoft}, which contains numerous real-world images with different exposure levels. 

\begin{figure}[t]
	\centering
	\includegraphics[width=1.0\linewidth]{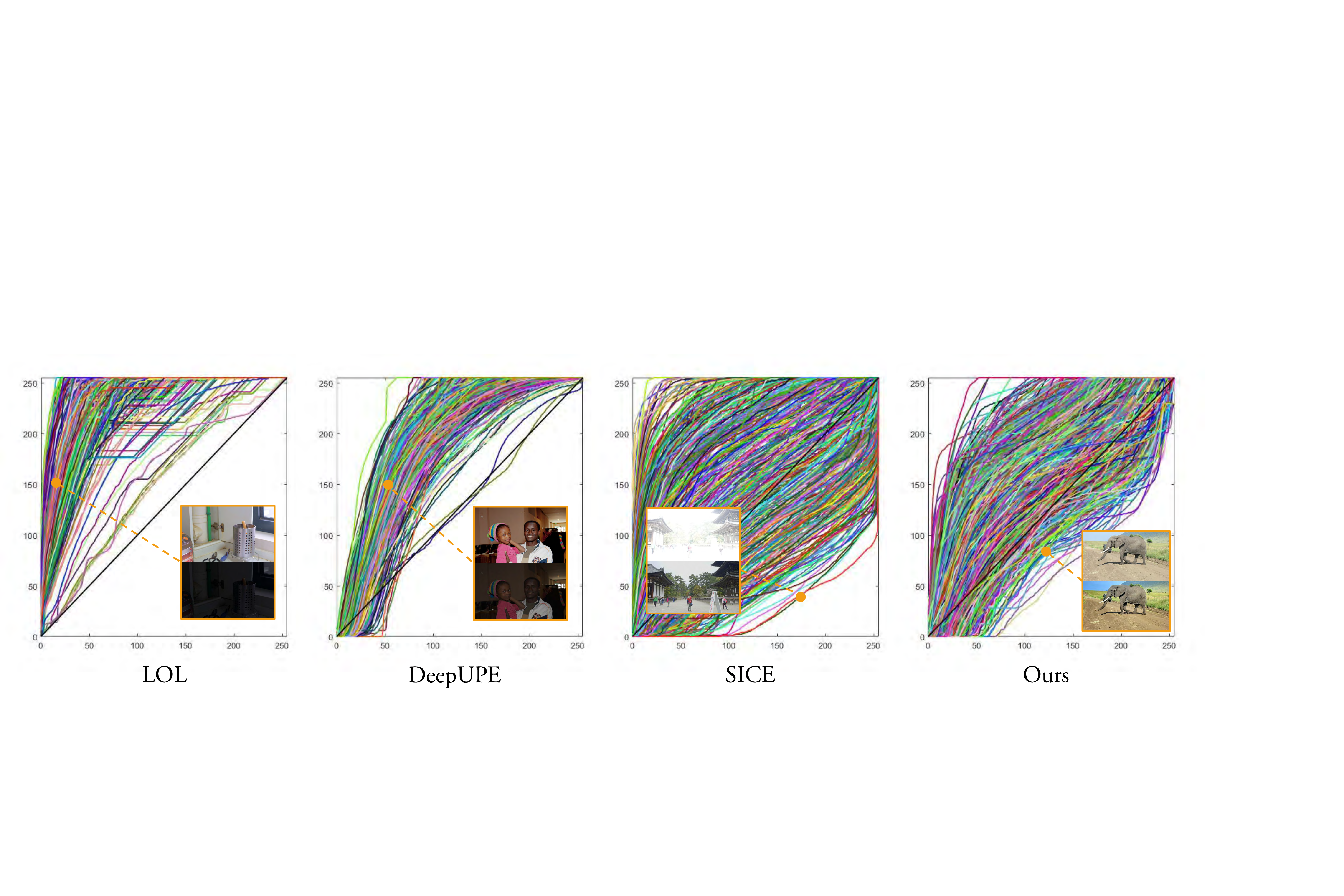}
	\caption{Existing representative enhancement datasets' and our dataset's statistical results of exposure adjustment curves. The small images are the example image pairs of different datasets.}
	\label{fig:compare}
\end{figure}

To visually show the differences between different enhancement datasets, we calculate the exposure adjustment curve, which is used to adjusting the histogram of the original images' Value component in HSV color space to match the histogram of reference images, as shown in Figure~\ref{fig:compare}. On one hand, the distribution of curves can approximately indicate the exposure adjustment of the dataset. That is to say, LOL~\cite{Chen2018Retinex} and DeepUPE~\cite{wang2019underexposed} are only used to learn to increase the exposure adaptively. SICE~\cite{Cai2018deep} and our dataset cover various exposure adjustments. 
On the other hand, the shape of the curve to some extent indicates the complexity of the adjustment. The curves of LOL~\cite{Chen2018Retinex} and SICE~\cite{Cai2018deep} are almost all simpler shapes similar to gamma curves, which shows that the covered light conditions are limited. As for DeepUPE~\cite{wang2019underexposed} and our dataset, the curve shapes are more complex similar to the S-Curve. As our dataset cover under-/over-exposed simultaneously, our light conditions are more diverse result in more complex curves compared with DeepUPE~\cite{wang2019underexposed}.
In summary, our dataset contains more diverse scenes and lighting conditions, which is a complement to existing datasets.

\subsection{Dataset Construction Details}
\label{sec:dataset_details}
The Microsoft COCO dataset~\cite{lin2014microsoft} covers diverse scenes, various resolution, different quality, manifold lighting conditions and abundant annotations, which is helpful for improving the robustness of the trained model. Therefore, we construct our new dataset based on COCO~\cite{lin2014microsoft}.
We design a semi-supervised retouching solution to automatically generate our dataset, instead of adjusting the images one by one using professional tools (like Photoshop). Specifically, we first cluster images based on their histograms. Then, images of the cluster center are selected and are adjusted using our retouching module to capture optimal coefficients according to human perception. Finally, according to the clustering results, the same coefficients are used for retouching images belong to the same class. 
In this experiment, we use the COCO train set ($\approx82$k images) and cluster this image empirically set to $500$ classes.

The key of our semi-supervised retouching solution is the retouching module, as shown in Figure~\ref{fig:simulation}. It can be formulated as:
\begin{equation}
	R_1\!=\!\mathcal{F}(I,\frac{I}{\mathcal{S}(max(I), \theta_1)^{\gamma_1}+\epsilon},1-\frac{1-I}{\mathcal{S}(max(1-I), \theta_2)^{\gamma_2}+\epsilon}, \theta_3)
\end{equation}
where $I$ and $R_1$ represent original image and the fusion result, $\epsilon$ is a small constant preventing division by zero, $\mathcal{S}$ and $\mathcal{F}$ represents smooth~\cite{xu2011image} and fusion~\cite{mertens2009exposure} operation, final result $R = R_1 + \alpha(R_1-\mathcal{S}(R_1, \theta_4))$, \{$\theta_1,\theta_2,\theta_3,\theta_4,\gamma_1,\gamma_2, \alpha$\} are the coefficients. 
Notice that, \{$\theta_1,\theta_2,\gamma_1,\gamma_2$\} are vectors to obtain image sequence with different enhancement. We first use the original/inverted image to enhance the under-/over-exposed regions to get preliminary enhancement sequence, and then fuse them and amplify the details to obtain the final satisfying image, inspired by~\cite{zhang2019dual,lv2019attention}.
The latent principle is enhancing contrast by locally smoothing the illumination and adjusting the exposure by gamma adjustment. 
Since our retouching solution is robust to similar light conditions (histograms), our semi-supervised retouching solution can efficiently enhance up to 82k images quickly.
Besides, we also simulate and add noise (using realistic noise model~\cite{Guo2019Cbdnet}) and compression artifacts (using JPEG compression) on the original COCO images, which are the two most common image degradation factors, to train our model for simultaneously suppressing noise and artifacts.

\begin{figure*}[htbp]
	\centering
	\includegraphics[width=0.98\textwidth]{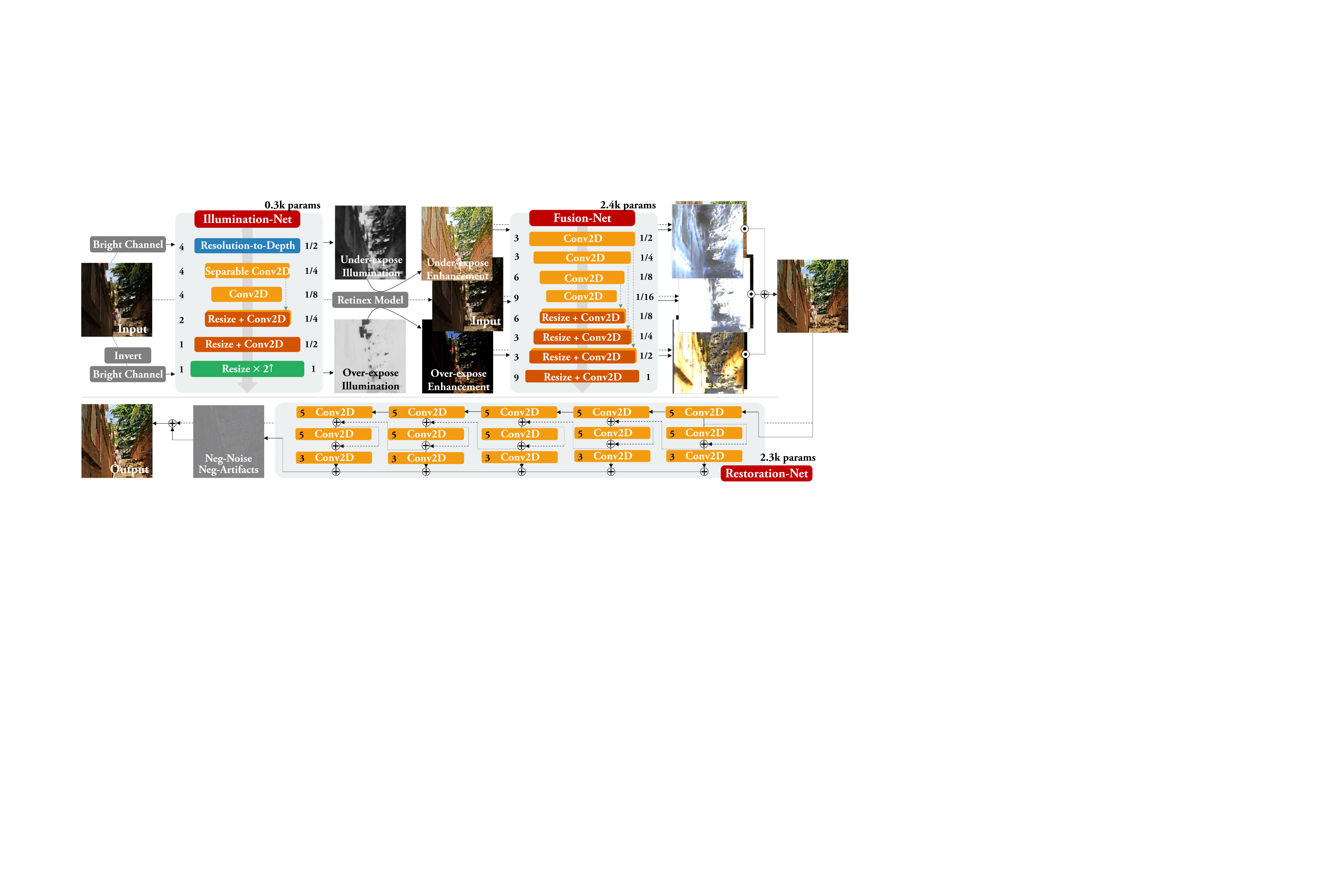}
	\caption{Overview of the proposed light-weight network architecture. The dashed lines represent skip connections. The Bright channel means the maximal value of three channels. $\odot$ and $\oplus$ represent pixel-wise multiplication and plus. The left and right numbers of every convolutional layer means the channel number and the resolution size compare with the input images.}
	\label{fig:network}
\end{figure*}

\section{Proposed Method}
In this section, we introduce the proposed solution, including enhancement model, network architecture, loss function and implementation details.

\subsection{Enhancement Model}
The Retinex model~\cite{land1977retinex} is a robust enhancement model, which aims to learn image-to-illumination instead of image-to-image mapping. 
The robust version~\cite{li2018structure} is formulated as: $R = I \circ L^{-1} + N$, where $I$, $L$ and $N$ represent original image, illumination map and negative noise map, $\circ$ denotes a pixel-wise multiplication. $R$ is the reflectance and usually used as the final enhancement result. 

\begin{figure}[htbp]
	\centering
	\includegraphics[width=1\linewidth]{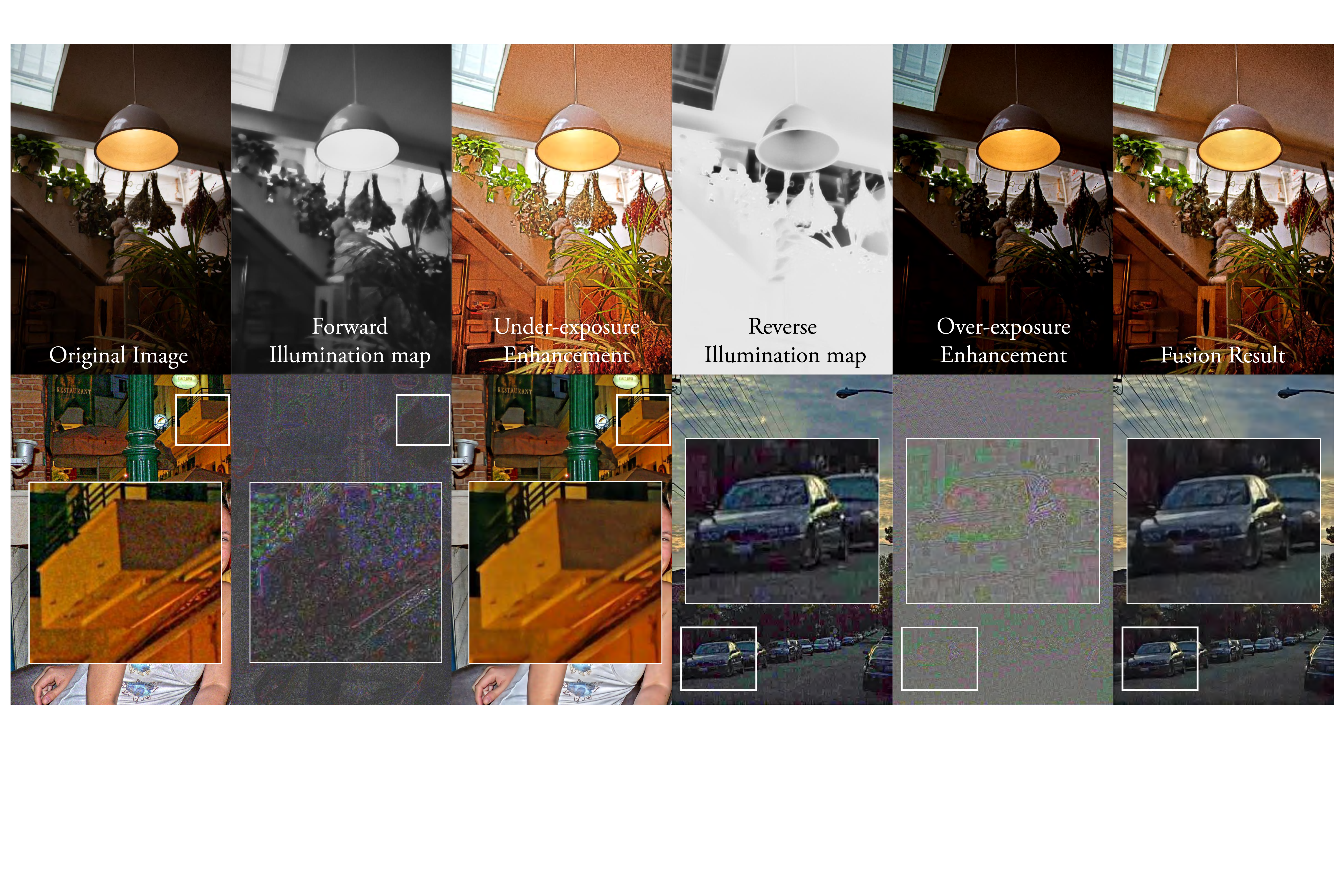}
	\caption{Examples of intermediate results of our model on real-world images. The noise map and artifacts map are normalized for better visualization.}
	\label{fig:mid_res}
\end{figure}

However, as the value range of the illumination map is $[0, 1]$, which means the prevalent Retinex-based enhancement models do not have the ability to suppress over-exposure regions of the non-uniform illumination images. Inspired by~\cite{zhang2019dual}, suppressing over-exposure regions of original images is equal to enhancing under-exposure regions of the inverted images. Thus, we can first enhance under-/over-exposure regions separately and then fusion them to generate final enhancement results (see figure~\ref{fig:mid_res}). In this way, we can keep the advantages (illumination maps have relatively simple forms with known priors for natural images) of the Retinex model and overcome its limitations (difficulty to suppress over-exposure regions).
The enhancement model can be formulated as:
\begin{equation}
	R = \mathcal{F}(I, I \circ L^{-1}, 1-(I_i \circ L_i^{-1}))+ N, 
\end{equation}
where $I_i$ and $L_i$ represent inverted image and the corresponding illumination map, $\mathcal{F}()$ represents the fusion function. 

\subsection{Network Architecture}
We propose a fully convolutional network containing three subnets: an Illumination-Net, a Fusion-Net and a Restoration-Net. Figure~\ref{fig:network} shows the overall network architecture.
As described in the enhancement model, the Illumination-Net is designed for estimating the illumination map based on the Retinex model. The Fusion-Net aims to fuse different intermediate enhanced results to generate exposure corrected images. The purpose of the Restoration-Net is to suppress the noise and compression artifacts. The detailed description is provided below.

{\bf Illumination-Net.} As the illumination is at least the maximal value of three channels at a certain location, we use the maximal value of three channels as the input of the Illumination-Net.
Also considering that the illumination maps have relatively simple forms with known priors for natural images, we can calculate the low-resolution illumination map and perform bilateral grid-based upsampling to enlarge the low-res prediction to approximate the full resolution illumination map~\cite{wang2019underexposed}. To avoid information loss caused by directly downsampling, we pack the input image into four channels and correspondingly reduce the spatial resolution by a factor of two in each dimension. 

{\bf Fusion-Net.} To better use the intermediate enhanced results, the output of the Fusion-Net is the fusion weight rather than the final fusion results. The final fusion result is formulated as: 
\begin{equation}
	R_1 = \mathcal{F}(I_U)\circ I_U+\mathcal{F}(I)\circ I+\mathcal{F}(I_O)\circ I_O, 
\end{equation}
where$R_1$ is the fusion result, $I$, $I_U$ and $I_O$ represent original image, under-expose enhancement result and the over-expose enhancement result, $\mathcal{F}()$ represents the Fusion-Net. We directly adopt U-Net in our implementation.

{\bf Restoration-Net.} According to the enhancement model, we design a light-weight multi-branch Restoration-Net to estimate the negative noise map $N$ to suppress the noise and compression artifacts, inspired by~\cite{lvmbllen}. Different from~\cite{lvmbllen}, we add skip connections between different branches to better reuse of extracted features. We directly calculate the sum of different branches' results as the final negative noise map.

\subsection{Loss Function}
We use a hierarchical strategy for training. Specifically, training is first done for Illumination-Net and Fusion-Net, which are as an end-to-end network. Then, training is done for Restoration-Net by fixing the weights of Illumination-Net and Fusion-Net. The detail loss functions of these two stages are given below.

{\bf Enhancement loss.} The training for Illumination-Net and Fusion-Net aims to improve the performance of enhancement, like contrast, colorfulness, detail, etc. To improve the image quality both qualitatively and quantitatively, we design a loss function by further  considering both structural and perceptual information. It can be expressed as:
\begin{equation}
	\mathcal{L}_E = \mathcal{L}_h + \mathcal{L}_p + \mathcal{L}_s + \omega_i\mathcal{L}_i,
\end{equation}
where the $\mathcal{L}_h$, $\mathcal{L}_p$, $\mathcal{L}_s$ and $\mathcal{L}_i$ represent Huber loss, 
structural loss, perceptual loss and illumination smoothness loss, and $\omega_i$ is the coefficient.

The Huber loss is a robust estimator and has proved to avoid the averaging problem of colorization~\cite{zhang2017real}. Similarly, it is useful for increasing the color saturation of images in enhancement tasks~\cite{Atoum2019color}. Therefore, we use Huber loss as the basic component of the loss function:
\begin{equation}
	\mathcal{L}_h = \frac{1}{2}(I_r-\widetilde{I})^2\mathbbm{1}_{\{|I_r-\widetilde{I}|<\delta\}} + \delta(|I_r-\widetilde{I}|-\frac{1}{2}\delta)\mathbbm{1}_{\{|I_r-\widetilde{I}|\ge\delta\}},
\end{equation}
where $I_r$ and $\widetilde{I}$ are the predicted and expected images. $\delta$ is the parameter of the Huber loss and is set to $0.5$ empirically. 

To reduce the perceptual error and improve the visual quality, we introduce perceptual loss by using VGG network~\cite{simonyan2014very} as the content extractor~\cite{ledig2016photo}. We use the output of the ReLU activation layers of the pre-trained VGG-19 network to define the perceptual loss as: 
\begin{equation}
	\mathcal{L}_{p} = \frac{1}{w_{ij}h_{ij}c_{ij}}\sum_{x=1}^{w_{ij}}\sum_{y=1}^{h_{ij}}\sum_{z=1}^{c_{ij}}
	\lVert\phi_{ij}(I_r)_{xyz}-\phi_{ij}(\widetilde{I})_{xyz}\lVert,
\end{equation}
where $w_{ij}$ , $h_{ij}$ and $c_{ij}$ describe the dimensions of the respective feature maps within the VGG-19 network. Besides, $\phi_{ij}$ indicates the feature map obtained by $j$-th convolution layer in $i$-th block of the VGG-19 Network.

The structural loss is introduced to preserve the image structure and avoid blurring and artifacts. We use the well-known image quality assessment algorithm SSIM~\cite{wang2004image} to estimate the structure error. It is defined as:
\begin{equation}
	\mathcal{L}_{s} = 1 - \frac{1}{N}\sum_{p\in img}\frac{2\mu_x \mu_y + C_1}{\mu_x^2+\mu_y^2+C_1}\cdot \frac{2\sigma_{xy} + C_2}{\sigma_x^2+\sigma_y^2+C_2},
\end{equation}
where $\mu_x$ and $\mu_y$ are pixel value averages, $\sigma_x^2$ and $\sigma_y^2$ are variances, $\sigma_{xy}$ is the covariance, and $C_1$ and $C_2$ are constants to prevent the denominator to zero.

Local consistency and structure-awareness are the key hypotheses for illumination estimation in previous works~\cite{guo2017lime,Chen2018Retinex,wang2019underexposed}. Following this idea, we introduce the illumination smoothness loss to smooth the textural details and preserve the overall structure boundary.
We use the structure-aware TV loss define the illumination smoothness loss as:
\begin{equation}
	\mathcal{L}_{i} = \Vert \nabla I_{i} \circ exp(-\lambda_g\cdot \nabla I) \Vert + \Vert \nabla \bar{I_{i}} \circ exp(-\lambda_g\cdot \nabla (1-I)) \Vert,
\end{equation}
where $I_{i}$ and $\bar{I_{i}}$ are the estimated forward and reverse illumination maps, $I$ is the original image, $\nabla$ represents the gradient, $\lambda_g$ is the coefficient balancing the strength of structure-awareness. We set $\lambda_g=10$ and $\omega_i=0.002$ empirically.

{\bf Restoration loss.} Image restoration also aims to preserve the structure, suppress noise and artifacts, and obtain satisfactory visual effects, which is the same as enhancement in some ways. Therefore, similar to the Enhancement loss, the Restoration loss is defined as:
\begin{equation}
	\mathcal{L}_E = \mathcal{L}_h + \mathcal{L}_p + \mathcal{L}_s + \omega_g\mathcal{L}_g,
\end{equation}
where the $\mathcal{L}_h$, $\mathcal{L}_p$ and $\mathcal{L}_s$ are the same as the corresponding components of the Enhancement loss.  $\mathcal{L}_g$ represents the global TV loss and is defined as $\Vert \nabla{I_r}\Vert$.
We empirically set $\omega_g=10^{-4}$ which denotes the coefficient of global TV loss.

\subsection{Implementation Details}
Our implementation is done with Keras~\cite{chollet2015keras} and Tensorflow~\cite{abadi2016tensorflow}.
The proposed light-weight network can be quickly converged after being trained for $10$ epochs on an Nvidia Titan Xp GPU using the proposed dataset. We use random clipping, flipping and rotating for data augmentation to prevent over-fitting. We set the batch-size to $32$ and the size of random clipping patches to $256\times256\times3$. We use the output of the fourth convolutional layer in the third block of the VGG-19 network~\cite{simonyan2014very} as the perceptual loss extraction layer. 
The input image values of Illumination-Net and Fusion-Net are scaled to $[0, 1]$, while the values are scaled to $[-1, 1]$ for Restoration-Net.
In the experiment, the entire network is optimized using the Adam optimizer~\cite{kingma2014adam} with parameters of $\alpha = 0.001$, $\beta_{1} = 0.9$, $\beta_{2} = 0.999$ and $\epsilon = 10^{-8}$. We also use the learning rate decay strategy, which reduces the learning rate to $98\%$ before the next epoch. At the same time, we reduce the learning rate to $50\%$ when the loss metric has stopped improving.

\section{Experimental Results}
In this section, we evaluate our method through extensive experiments. We first compare our method with state-of-the-art enhancement methods in both qualitative and quantitative. Then, we present more analysis to demonstrate our method comprehensively.

\subsection{Comparison with State-of-the-art Methods}
We comprehensively compare our method with state-of-the-art methods by using the publicly-available codes with recommended parameter settings to show that our method is complementary to existing methods. 

\begin{figure}[t]
	\centering
	\includegraphics[width=1\linewidth]{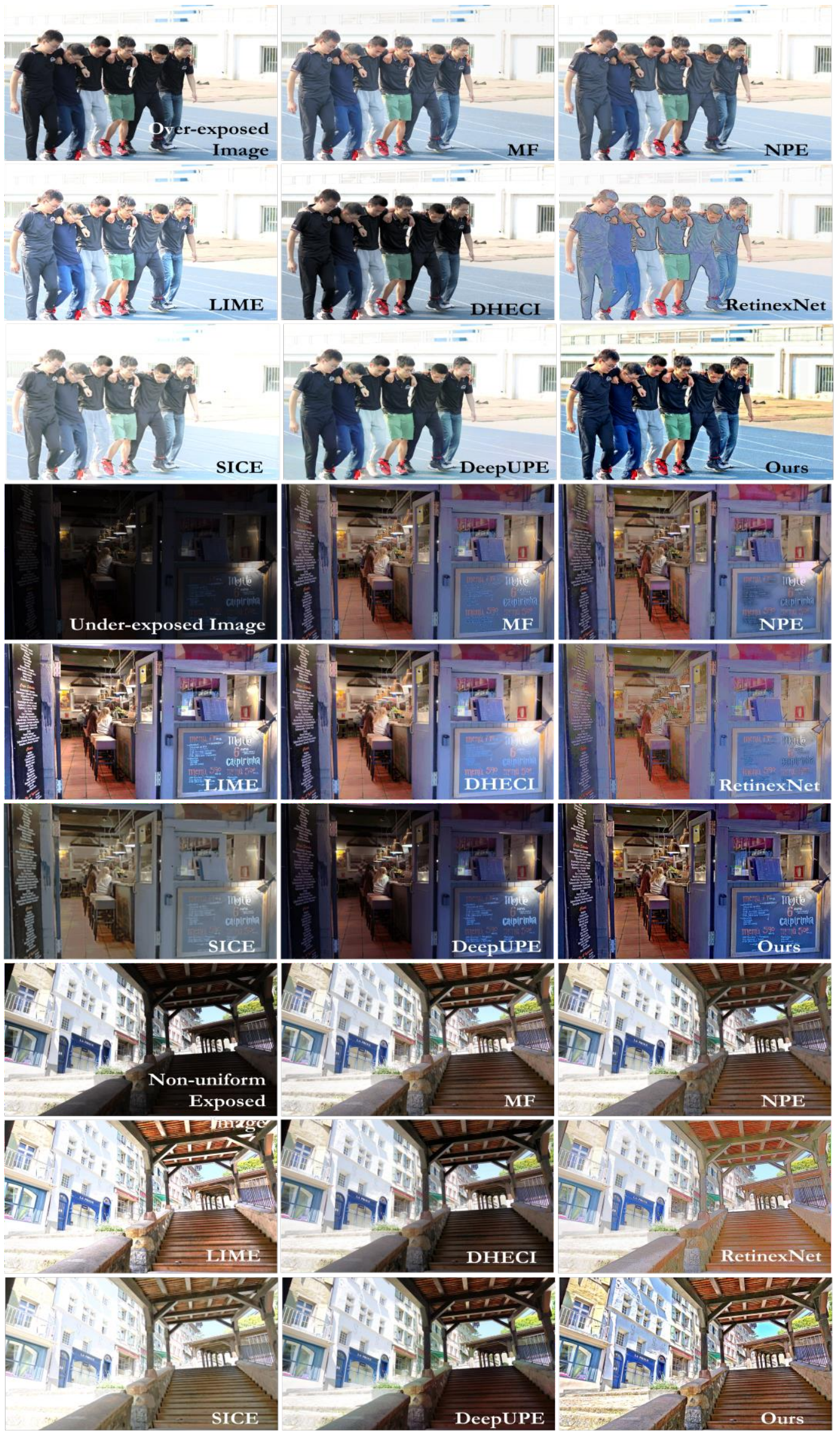}
	\caption{Visual comparison of real-world challenging non-uniform exposed images. Please zoom in for a better view.}
	\label{fig:ex1}
\end{figure}

{\bf Visual Comparison.} We provide a visual comparison to show the differences between our method and existing state-of-the-art algorithms. Typical challenging cases are shown in Figure~\ref{fig:ex1}. 

For the first over-exposed scene, enhancing dark clothes is challenging as they are easily confused with under-exposed regions. This dilemma is especially serious for Retinex-based methods, like RetinexNet~\cite{Chen2018Retinex} and LIME~\cite{guo2017lime}. Our method can avoid this problem to some extent by image fusion strategy. Besides, for over-exposed regions like runways and stands, these methods fail to enhance them. In contrast, our method effectively enhances over-exposed regions and obtain high contrast and rich color.

For the second under-exposed scene, insufficient enhancement (like DeepUPE~\cite{wang2019underexposed}), color degradation (like SICE~\cite{Cai2018deep}), and local over-enhancement (see regions of the light source in NPE~\cite{wang2013naturalness}) are flaws of existing methods. In contrast, our method is able to reveal vivid colors, avoid over-/under- enhancement, and improve the details simultaneously. 

For the last scene, over-/under-exposed regions need to be enhanced simultaneously. Existing methods tend to enhance under-exposed regions but ignore the over-exposed ones. Our method effectively enhances different exposed regions simultaneously and amplifies the contrast, which makes results more appealing.

In addition, our method is able to enhance the 720p video frame-by-frame almost in real-time. Our method also outperforms these methods on video enhancement. Please check the supplementary materials for details.

{\bf Quantitative Comparison.} To evaluate the inference performance and generalization capability of our solution, we quantitatively compare it with the other methods. For a fair comparison of generalization capability, we build a test set contains 50 various exposed images selected from existing paired public enhancement datasets (15 images from LOL~\cite{Chen2018Retinex}, 15 images from SICE~\cite{Cai2018deep} and 20 images from DeepUPE~\cite{wang2019underexposed}).
Tables~\ref{table:ex1} reports the comparison results, where for every method, we use the pre-trained weights or recommended parameters. Our result performances well in all quality metrics, which fully demonstrates the outperformance of our approach. 

\begin{table*}
	\begin{center}
		\caption{Quantitative comparison results. The average runtime is tested using images with size $1280\times720$. ``*" represents only using an Intel i5-8400 CPU.}
		\label{table:ex1}
		\begin{tabular}{l|ccccc|cc}
			\hline
			Algorithm & $\uparrow$PSNR & $\uparrow$SSIM~\cite{wang2004image} &$\uparrow$VIF~\cite{sheikh2006image} &$\downarrow$LOE~\cite{ying2017bio} & $\downarrow$NIQE\cite{mittal2012making}  & Params & Runtime\\
			\hline\hline
			*MSR~\cite{jobson1997multiscale} & 11.87 & 0.56 & \cellcolor{lvorange}{0.41} & 2029.4 & 4.19 & - & 1.44s\\
			*Dong~\cite{dong2011fast} & 13.82 & 0.54 & 0.33 & 1598.0 & 4.91 & - & 0.43s\\
			*BPDHE~\cite{ibrahim2007bpdhe} & 14.41 & 0.57 & 0.34 & 892.2 & 4.21 & - & 0.49s\\
			*NPE~\cite{wang2013naturalness} & 14.95 & 0.58 & 0.38 & 1563.7 & 4.31 & - & 25.6s\\
			*DHECI~\cite{nakai2013dheci} & 16.14 & 0.58 & 0.39 & 903.3 & 4.62 & - & 42.3s\\
			*MF~\cite{fu2016mf} & 16.10 & 0.62 & 0.39 & 1113.1 & 4.51 & - & 0.83s\\
			*LIME~\cite{guo2017lime} & 12.49 & 0.53 & \cellcolor{lvred}{0.42} & 1441.2 & 4.68 & - & 0.56s\\
			*BIMEF~\cite{ying2017bio} & 15.58 & 0.66 & \cellcolor{lvyellow}{0.40} & \cellcolor{lvred}{857.1} & 3.97 & - & 0.54s\\
			\hline
			SICE~\cite{Cai2018deep} & 14.63 & 0.62 & 0.31 & 1312.2 & 4.24 & 682k & 1.81s\\
			RetinexNet~\cite{Chen2018Retinex} & 12.84 & 0.51 & 0.31 & 2278.2 & 5.07 & \cellcolor{lvyellow}{445k} & \cellcolor{lvyellow}{0.16s}\\
			GLADNet~\cite{wang2018gladnet} & \cellcolor{lvyellow}{17.71} & \cellcolor{lvyellow}{0.68} & 0.36 & 949.9 & 3.87 & 932k & 0.38s\\
			MBLLEN~\cite{lvmbllen} & \cellcolor{lvred}{18.06} & \cellcolor{lvorange}{0.71} & 0.33 & 898.1 & \cellcolor{lvorange}{3.06} & 450k & 0.31s\\
			DeepUPE~\cite{wang2019underexposed} & 16.48 & 0.65 & \cellcolor{lvyellow}{0.40} & \cellcolor{lvyellow}{871.4} & \cellcolor{lvyellow}{3.69} & \cellcolor{lvorange}{100k} & \cellcolor{lvorange}{0.10s}\\
			Ours & \cellcolor{lvorange}{17.83} & \cellcolor{lvred}{0.73} & \cellcolor{lvred}{0.42} & \cellcolor{lvorange}{869.7} & \cellcolor{lvred}{3.03} & \cellcolor{lvred}{5k} & \cellcolor{lvred}{0.05s}\\ \hline
		\end{tabular}
	\end{center}
\end{table*}

For inference performance, our method significantly outperforms other methods. Our model is very lightweight, which makes it potentially useful for mobile devices. Besides, the inference speed of our model is very fast. It can enhance 0.5 mega-pixel images in real-time and 720p video in almost real-time (20f/s).

\begin{figure}[t]
	\centering
	\includegraphics[width=1\linewidth]{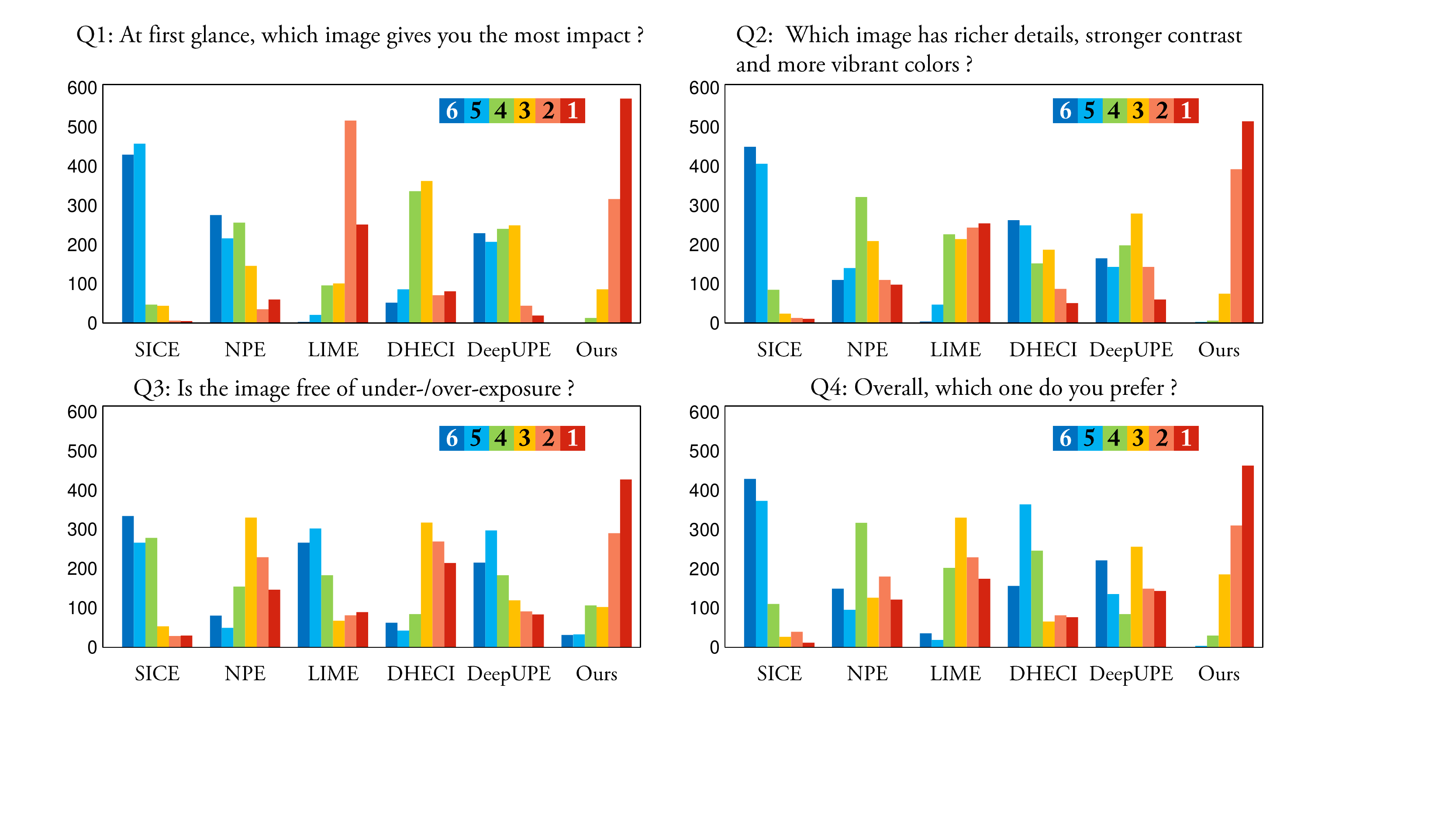}
	\caption{Rating distributions on four questions of our user study. The ordinate axis is the ranking frequency received by the methods from the participants.}
	\label{fig:user}
\end{figure}

{\bf User Study.} To test the subjective preference of non-uniform exposed image enhancement methods, we conduct a user study with $50$ participants. We randomly select $20$ natural non-uniform exposed images and enhance them using our method and other five representative methods. For each case, the original image and six enhanced results are displayed to the participants simultaneously in a random arrangement. Then, the participants are asked to rank the quality of the six enhancements from 1 (best) to 6 (worst) for each of the four questions.
We also provide zoom-in function to let participants check details. 
Figure~\ref{fig:user} shows the statistical result of the user study, where every sub-figure summarizes the rating distribution of a particular question. Our method receives more ``best" ratings, which shows that our method is more preferred by human subjects.

\begin{figure}[h]
	\centering
	\includegraphics[width=1\linewidth]{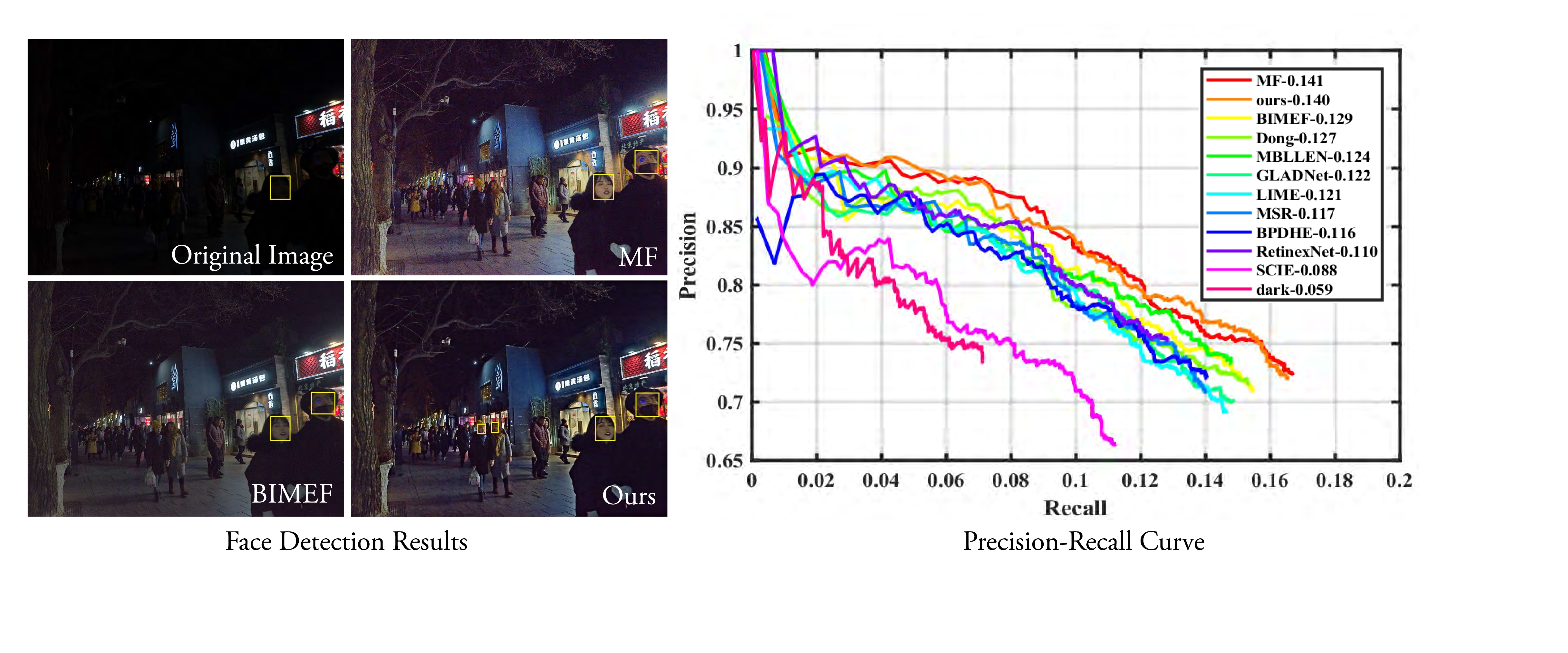}
	\caption{Face detection result comparison. {\bf Left}: An intuitive example of visual comparison. {\bf Right}: precision-recall curves and average precisions of after enhanced using different methods. ``dark" means the result of original images without any pre-processing.}
	\label{fig:face}
\end{figure}

{\bf Face Detection at night.} Image enhancement aims to improve visibility and reflect clear details of target scenes, which are critical to many vision-based techniques especially under poor conditions. We take face detection at night as an example to investigate the effects of different enhancement methods for improving detection performance. We use the DARK FACE dataset~\cite{yuan2019ug} for testing, which contains 10k low-light images with corresponding face annotation. 
We use the pre-trained light-weight version\footnote{https://github.com/lijiannuist/lightDSFD} of DSFD~\cite{li2019dsfd}, which is the state-of-the-art deep face detector, to investigate the performance of real-time detection.
To clearly demonstrate the gap between different enhancement algorithms, we select 500 ``easy" images for evaluation by using the DARK FACE evaluation tool\footnote{https://flyywh.github.io/CVPRW2019LowLight/}. 
The comparison of precision-recall (P-R) curves and the average precisions (AP) are shown in Figure~\ref{fig:face}.
All these enhancement methods are beneficial to improve detection performance. Among these methods, our method and MF~\cite{fu2016mf} perform best, which means to some extent that our results can effectively and realistically reflect the details of real scenes. Besides, compared with MF~\cite{fu2016mf}, our method is faster and can be trained together with face detectors which means that our method is more appealing in real applications.

\subsection{More Analysis}
We provide more analysis to explore the role of components of our model and discuss the flexibility, extendibility and limitation of our method.

\begin{figure}[t]
	\begin{center}
		\begin{overpic}[width=1\linewidth]{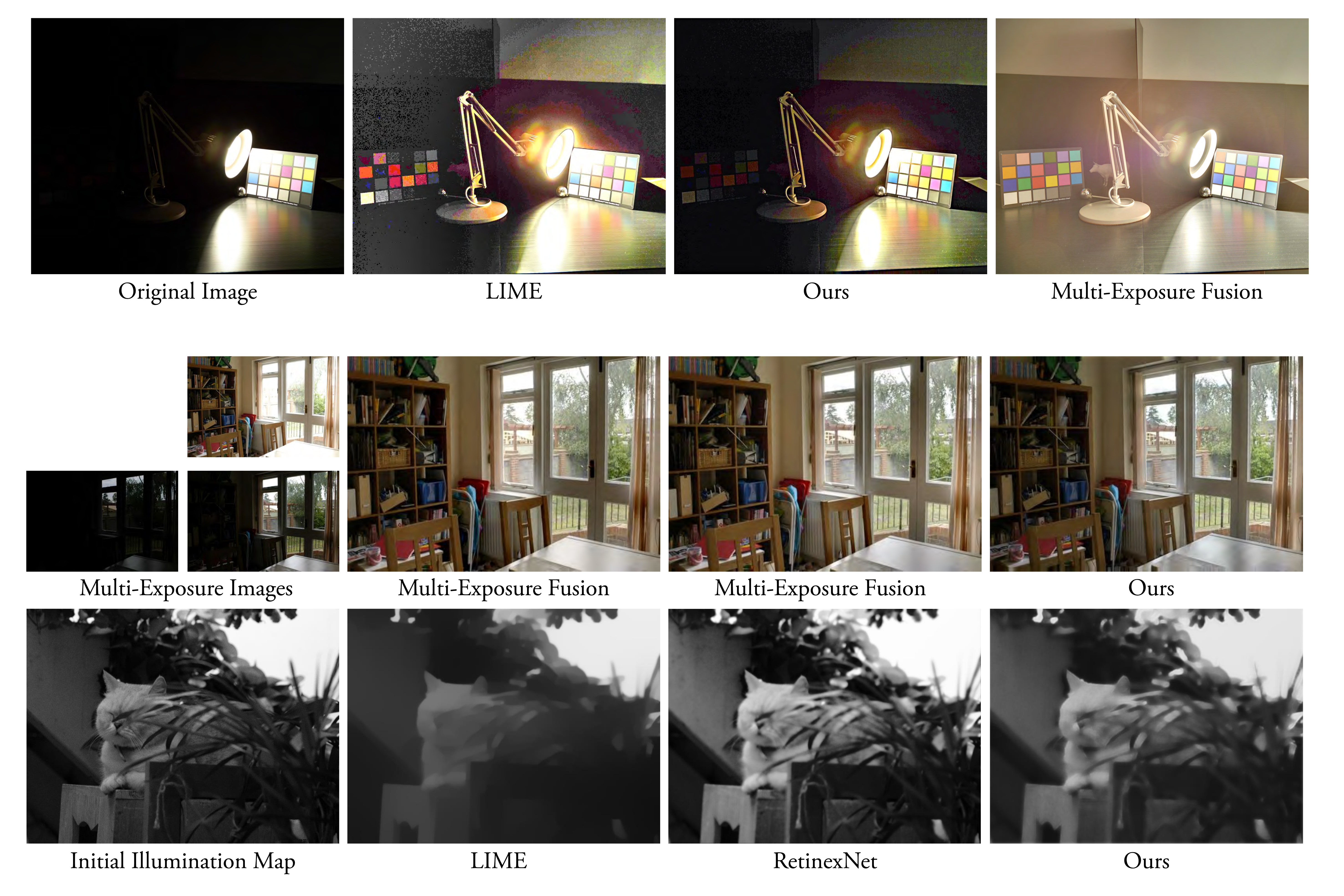}            
			\put(111,53){\bf \color{black}\tiny \cite{kou2017multi}} 
			\put(170,53){\bf \color{black}\tiny \cite{mertens2009exposure}} 
		\end{overpic}
	\end{center}
	\caption{{\bf Top}: Multi-Exposure Fusion comparison with MEF methods. {\bf Bottom}: Illumination maps estimation comparison with typical Retinex-based methods.}
	\label{fig:why}
\end{figure}

{\bf Why our Model Works?} 
As illumination maps of natural images typically have relatively simple forms with known priors, learning an image-to-illumination mapping is easier than image-to-image mapping on photographic adjustment under diverse lighting conditions~\cite{wang2019underexposed}.
Hence, our Illumination-Net has ability to customizing the inputs (like adjusting exposure and contrast) to the Fusion-Net by formulating constraints (like adjusting illumination magnitudes and enforcing locally smooth) on the estimated illumination map, as shown in Figre~\ref{fig:why}.
However, according to the Retinex model, using a single illumination map fails to enhance both under-/over-exposed areas simultaneously.

Therefore, to overcome this dilemma, we introduce two illumination maps for enhancing under-exposure and suppressing over-exposure respectively and fuse them using Fusion-Net by estimating the fusion weight map. The final results can be customized by adjusting the fusion weight map, which provides stronger generalization capabilities and learning capabilities for our model to learn complex adjustment for both under-exposure and over-exposure regions simultaneously.

To demonstrate the good generalization capability of our network, we directly fuse real multi-exposure images using our Fusion-Net without any fine-tuning. Our fusion result is comparable with the latest fusion methods as shown in Figure~\ref{fig:why}, which shows the good adaptability and robustness of our network.
In summary, our model has strong generalization and learning capabilities to learn and enhance non-uniform exposed images adaptively.

\begin{figure}[t]
	\centering
	\includegraphics[width=1\linewidth]{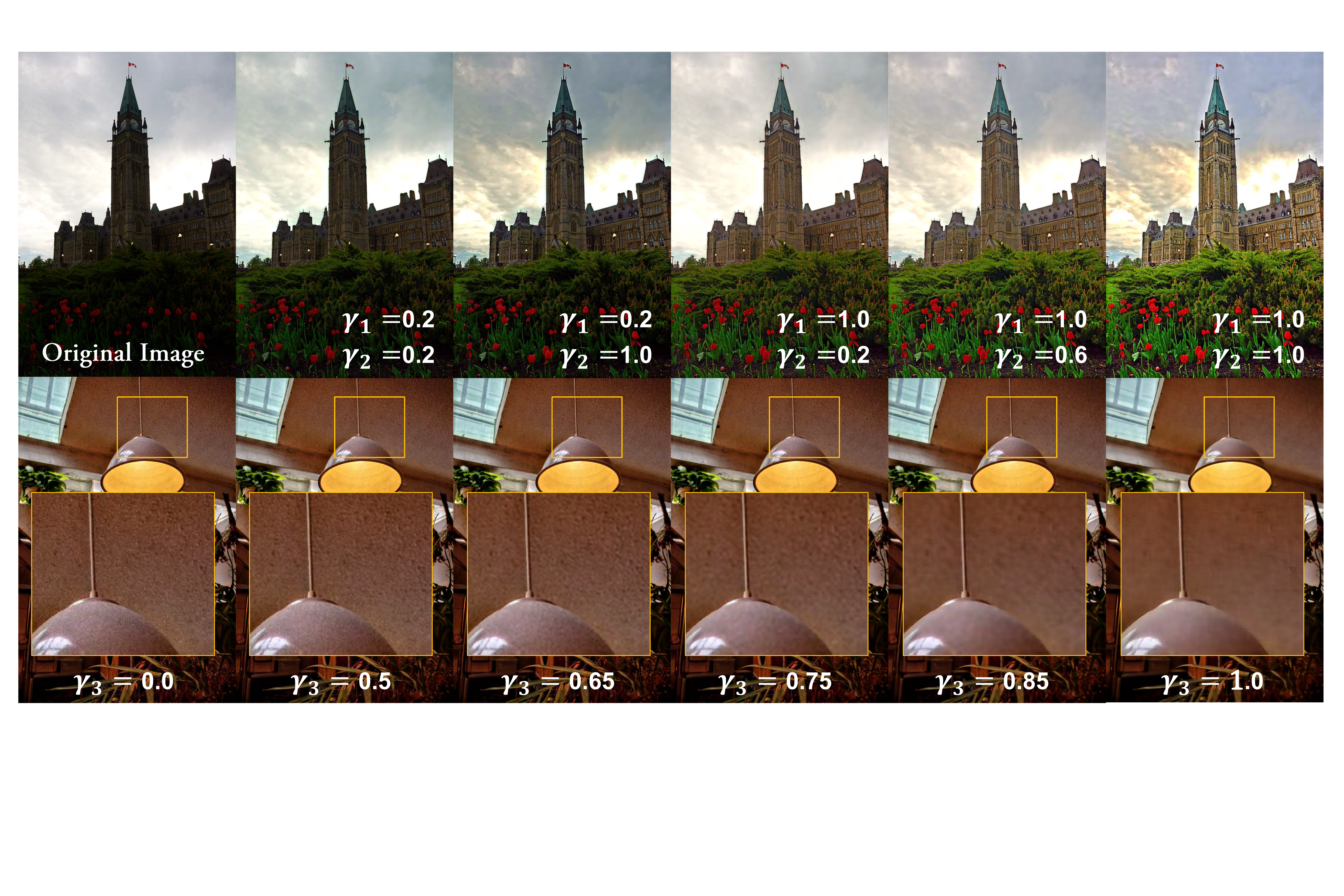}
	\caption{Examples of interactive enhancement and interactive noise removal.}
	\label{fig:iteration}
\end{figure}

{\bf Interactive Enhancement.} Considering that the assessment of enhancement results is subjective, providing interactive enhancement is necessary for some application scenes. We formulate the interactive enhancement model as:
\begin{align}
	R\!=\!\mathcal{F}(I, I/({L}^{\gamma_1}+\epsilon), 1-I_i/({L_i}^{\gamma_2}+\epsilon)) + \mathcal{D}^{-1}(\Phi(\mathcal{D}(N), \gamma_3)),
\end{align}
where $I$, $L$, $N$ and $R$ represent original image, illumination map, estimated negative noise and the final interactive results, $I_i$ and $L_i$ represent inverted image and the correspond illumination map, $\epsilon$ is a small constant preventing division by zero. $\mathcal{F}$ represents fusion operation, $\mathcal{D}$ and $\mathcal{D}^{-1}$ are discrete cosine transform (DCT) and the inverse transform, $\Phi()$ represents retaining high-frequency components and setting others to zero.
$\gamma_1$,$\gamma_2$ and $\gamma_3$ are the interactive coefficients, which control enhancing under-exposed regions, suppressing over-exposed areas and noise removal, respectively. 
We set the value range of $\gamma_1$,$\gamma_2$ and $\gamma_3$ to $[0,1]$. The larger value of $\gamma_1$ ($\gamma_2$), the stronger enhancement of under-exposed (over-exposed) regions, as shown in Figure~\ref{fig:iteration}. Similarly, larger value of $\gamma_3$ means more noise are removed. Proper $\gamma_3$ makes a trade-off between denoising and texture retaining.

\begin{figure}[t]
	\centering
	\includegraphics[width=1\linewidth]{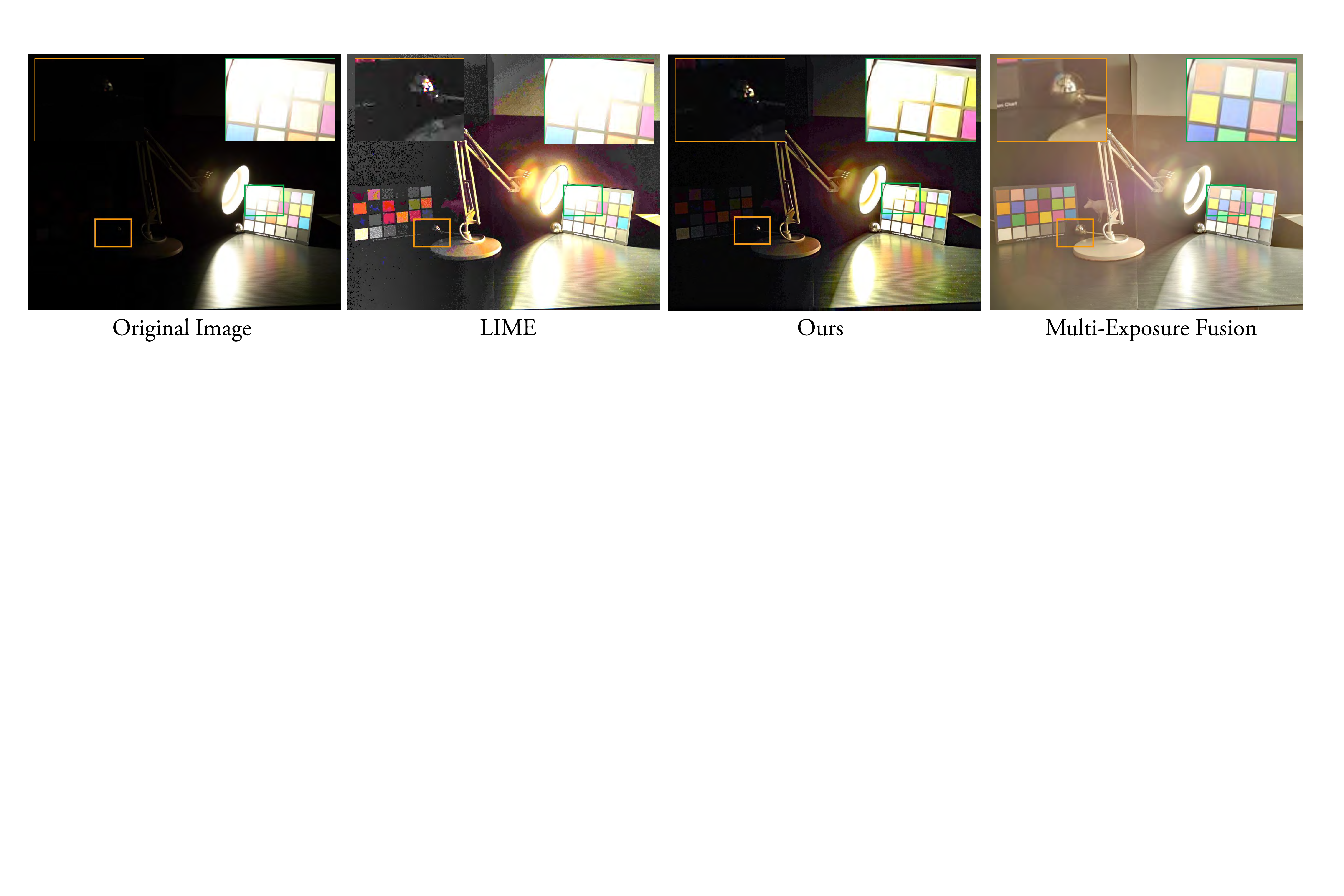}
	\caption{A Failure case. The missing details can be found in the MEF results (using 18 images for fusion).  }
	\label{fig:failure}
\end{figure}

{\bf Ablation Study.}
We quantitatively evaluate the effectiveness of different components in our method based on our proposed dataset using PSNR and SSIM~\cite{wang2004image} as the metrics, as shown in Table~\ref{table:ablation}.
Note that the Restoration-Net is not considered in this study. Directly learning image-to-image mapping using light-weight network will severely reduce enhancement quality (condition 1-2), which shows the effectiveness of our network architecture. We use $mse$ as the naive loss function under condition 2. The results (condition 3-6) show that the quality of enhancement is improving by containing more loss components. For the effect of model size, larger models bring little gain (especially for visual perception), but lighter networks reduce the quality severely (condition 7-8).

\begin{table}
	\begin{center}
		\caption{Ablation study. This table reports the performance under each condition based on our proposed dataset. In this table, "w/o" means without.}
		\label{table:ablation}
		\begin{tabular}{l|cccc}
			\hline
			~~Condition  & ~~~PSNR~~~& ~~~SSIM~~~ \\ \hline\hline
			~~1. U-Net ($\approx3.0$k params) &  18.63 & 0.78 \\
			~~2. cGAN ($\approx3.0$k params) &  17.46 & 0.71 \\
			\hline
			~~3. w/o $\mathcal{L}_{h}$, w/o $\mathcal{L}_{s}$, w/o $\mathcal{L}_{p}$, w/o $\mathcal{L}_{i}$ &  20.26 & 0.87 \\
			~~4. with $\mathcal{L}_{h}$, w/o $\mathcal{L}_{s}$, w/o $\mathcal{L}_{p}$, w/o $\mathcal{L}_{i}$ &  21.01 & 0.86 \\
			~~5. with $\mathcal{L}_{h}$, with $\mathcal{L}_{s}$, w/o $\mathcal{L}_{p}$, w/o $\mathcal{L}_{i}$ &  20.92 & 0.90 \\
			~~6. with $\mathcal{L}_{h}$, with $\mathcal{L}_{s}$, with $\mathcal{L}_{p}$, w/o $\mathcal{L}_{i}$ &  21.85 & 0.90 \\ 
			\hline
			~~7. Dwindling model ($\approx1.5$k params) &  20.06 & 0.87 \\
			~~8. Enlarging model ($\approx9.1$k params) &  {\bf 22.89} & 0.91 \\
			\hline\hline
			~~9. Proposed ($\approx2.7$k params) &  22.68 & {\bf 0.92} \\
			\hline
		\end{tabular}
	\end{center}
\end{table}

{\bf Limitation.} Our method can produce satisfactory results for most non-uniform exposed images as validated above. However, for those regions without any trace of texture (complete under-exposure or over-exposure), our method fails to recover the details. Figure~\ref{fig:failure} presents an example case where our method, as well as other state-of-the-art methods, all fail to produce satisfying results.

\section{Conclusions}
We propose an end-to-end light-weight network for non-uniform illumination image enhancement. Different from Retinex-based methods, our method can suppress over-exposure regions by enhancing under-exposure regions of the inverted version, which keeps the advantages (illumination maps have relatively simple forms with known priors) of the Retinex model and overcome its limitations (unable to enhance over-/under-exposure regions simultaneously). We also propose a semi-supervised retouching solution to construct a new dataset ($\approx82$k image pairs) for our network to handle color, exposure, contrast, noise and artifacts, etc., simultaneously and effectively.
Extensive experiments demonstrate the effectiveness of our model. Our network only has $5000$ parameters and can enhance 0.5 mega-pixel images in real-time ($\approx50$ fps), which is faster than existing enhancement algorithms.

Our future work will focus on recovering the missing image content for extremely under-exposed or over-exposed regions (see Figure~\ref{fig:failure}) by using semantics information guided or texture synthesis techniques. 

\bibliographystyle{ACM-Reference-Format}
\bibliography{sample-base}

\end{document}